\pdfoutput=1
\documentclass[11pt]{article}

\usepackage{acl}

\usepackage{times}
\usepackage{latexsym}

\usepackage[T1]{fontenc}
\usepackage[utf8]{inputenc}

\usepackage{microtype}

\usepackage{inconsolata}

\usepackage{hyperref}
\usepackage{url}
\usepackage{diagbox}
\usepackage{algorithm}
\usepackage{algpseudocode}
\usepackage{amsmath}
\usepackage{amssymb}
\usepackage{amsmath}
\usepackage{arydshln}
\usepackage{ascii}
\usepackage{bm}
\usepackage{booktabs}
\usepackage{colortbl}
\usepackage{color}
\usepackage{CJK}
\usepackage{dsfont}
\usepackage{enumitem}
\usepackage{forest}
\usepackage{graphics}
\usepackage{graphicx}
\usepackage{latexsym}
\usepackage{multirow}
\usepackage{mwe}
\usepackage{makecell}
\usepackage{microtype}
\usepackage{subfig}
\usepackage{times}
\usepackage{tikz}
\usepackage{tabularx}
\usepackage{url}
\usepackage{wrapfig}
\usepackage{tcolorbox}
\usepackage{xcolor}
\definecolor{comments}{RGB}{0, 150, 0}

\definecolor{highlight}{RGB}{200, 0, 0}

\definecolor{sred}{RGB}{203, 64, 46}
\definecolor{sblue}{RGB}{44, 73, 135}
\definecolor{sgreen}{RGB}{37, 100, 28}

\title{LLM$\times$MapReduce: Simplified Long-Sequence Processing\\using Large Language Models}

\author{Zihan Zhou$^{1*}$, Chong Li$^{2*}$, Xinyi Chen$^{3}$\thanks{ Equal Contribution.
}, Shuo Wang$^{4\dag}$, Yu Chao$^{4}$,\\
\textbf{Zhili Li}$^{5}$\textbf{,}
\textbf{Haoyu Wang}$^{5}$\textbf{,}
\textbf{Rongqiao An}$^{4}$\textbf{,}
\textbf{Qi Shi}$^{4}$\textbf{,}
\textbf{Zhixing Tan}$^{4}$\textbf{,}\\
\textbf{Xu Han}$^{4,6,7}$\textbf{,}
\textbf{Xiaodong Shi}$^{1\dag}$\textbf{,}
\textbf{Zhiyuan Liu}$^{4,6,7}$\thanks{ Correspondence to Shuo Wang, Xiaodong Shi, and Zhiyuan Liu.}\textbf{,}
\textbf{Maosong Sun}$^{4,6,7}$\\
$^{1}$Xiamen University
$^{2}$Peking University
$^{3}$Nankai University\\
$^{4}$Dept. of Comp. Sci. \& Tech., Tsinghua University
$^{5}$BUPT\\
$^{6}$Institute for AI, Tsinghua University \\
$^{7}$Beijing National Research Center for Information Science and Technology
}

\begin{document}
\maketitle
\begin{abstract}
Enlarging the context window of large language models (LLMs) has become a crucial research area, particularly for applications involving extremely long texts.
In this work, we propose a novel training-free framework for processing long texts, utilizing a divide-and-conquer strategy to achieve comprehensive document understanding.
The proposed LLM$\times$MapReduce framework splits the entire document into several chunks for LLMs to read and then aggregates the intermediate answers to produce the final output. The main challenge for divide-and-conquer long text processing frameworks lies in the risk of losing essential long-range information when splitting the document, which can lead the model to produce incomplete or incorrect answers based on the segmented texts.
Disrupted long-range information can be classified into two categories: {inter-chunk dependency} and {inter-chunk conflict}.
We design a {\em structured information protocol} to better cope with inter-chunk dependency and an {\em in-context confidence calibration} mechanism to resolve inter-chunk conflicts. Experimental results demonstrate that LLM$\times$MapReduce can outperform representative open-source and commercial long-context LLMs, and is applicable to several different models.\footnote{ The code is available at \url{https://github.com/thunlp/LLMxMapReduce}.}
\end{abstract}

\section{Introduction}
\label{sec:intro}
Large language models (LLMs) exhibit impressive performance across a wide range of complex tasks~\citep{gpt4}, including question answering~\citep{claude2}, code generation~\citep{luo2024wizardcoder}, and solving mathematical problems~\citep{luo2023wizardmath}. However, due to their quadratic computational complexity and a lack of high-quality long training examples, most LLMs are trained with a limited window size~\citep{llama,llama2,jiang2023mistral}. This context limit restricts the application of modern LLMs to long-sequence processing tasks. In response to this issue, several researchers have focused on extending the context length of LLMs. Existing studies can be broadly categorized into two types: training-based and training-free methods.

For training-based extension methods, it is necessary to prepare long training data and allocate substantial computational resources to support the additional training.
% \comments{Add training-based works: LlamaLong (ABF), LongLoRA, Data Engineering (Yao Fu), LongAlign, DeepSeek-V2, Qwen2, Llama3.1.}
\citet{xiong2023effectivelongcontextscalingfoundation} propose adjusting the base frequency of RoPE~\citep{su2023roformerenhancedtransformerrotary}, and then training the model with 400 billion tokens of long text, significantly improving its performance on long-context tasks.
\citet{chen2024longloraefficientfinetuninglongcontext} introduce LongLoRA, which employs shifted sparse attention for efficient fine-tuning and utilizes learnable embedding and normalization layers during long-context fine-tuning.
Although these training-based methods can effectively extend the context length of LLMs, they may be inapplicable in scenarios where sufficient computational resources and high-quality long texts are unavailable.

By contrast, training-free context extension approaches aim to break the length limit of LLMs without tuning their parameters. 
% \comments{Add training-free works: YaRN, Dual-chunk attention, head retrieval (Fudan), Inf-LLM, Steaming-LLM.}
For example, \citet{xiao2024efficientstreaminglanguagemodels} suggest preserving several initial tokens within a sliding window attention mechanism, thereby enabling large language models to process unlimited text without the need for fine-tuning.
InfLLM~\citep{xiao2024infllmtrainingfreelongcontextextrapolation}, also leveraging sliding window attention, employs additional memory units to store distant contexts and incorporates an efficient mechanism for retrieving relevant historical information.
Another prominent research direction employs the divide-and-conquer idea, processing long sequences by splitting them into shorter chunks.
% \comments{ICLR work, LongAgent, Huawei work, Qwen Agent, Chain-of-Agents, LC-Boost}
LangChain~\citep{langchain2022} initially introduces the MapReduce method, where text segments are processed in parallel during the map stage, followed by the aggregation of intermediate results across all segments to predict the final output. Similarly, in XL$^3$M~\citep{wang2024xl3mtrainingfreeframeworkllm}, long texts are divided into multiple short sub-contexts, each paired with a question. Relevant segments are then selected using LLMs and combined chronologically to generate the final answer.
The major challenge for this kind of method is that different segments are processed independently, which may break some essential long-range information.
Disrupted long-range information can be divided into two categories: (1) {\em inter-chunk dependency}, where evidence is spread across different chunks and relies on each other; and (2) {\em inter-chunk conflict}, where evidence across chunks is contradictory, requiring the model to resolve these conflicts in order to predict the final answer.
 
% ~\cite{qwen-agent-2405} developed an agent with three levels of complexity, each building upon the previous one. \comments{Level 1 uses Retrieval-Augmented Generation (RAG). Level 2 processes all chunks, outputs relevant sentences, uses them as search queries to find the most relevant chunks within an 8k-context limit using BM25, and then generates the final answer. Level 3 involves step-by-step reasoning, where the agent decomposes questions into sub-questions, has the Level 2 agent answer them, updates its memory, and combines the answers to produce the final response.}
To address the challenges of inter-chunk dependency and inter-chunk conflict, recent works have proposed more advanced divide-and-conquer frameworks.
LongAgent~\citep{zhao2024longagentscalinglanguagemodels} introduces a framework comprising a leader agent and multiple member agents, each responsible for processing a chunk, all powered by LLMs. Each member provides an answer to the leader, who groups the responses and randomly selects a representative from each group to determine the final answer. However, our experiments show that LongAgent's aggregation mechanism does not effectively resolve inter-chunk dependency and conflict, as randomly selecting members can result in the loss of important evidence. Unlike LongAgent, which processes multiple chunks in parallel, Chain-of-Agents (CoA)~\citep{zhang2024chainagentslargelanguage} sequentially processes split chunks using an accumulated summary. However, because CoA's workflow does not explicitly address the inter-chunk conflict problem, it performs worse than LongAgent in our experiments.
LC-Boost~\citep{qian2024longllmsnecessitylongcontexttasks} defines an action space and selects appropriate actions for sequentially processing chunks. To address inter-chunk conflicts, LC-Boost adaptively either appends new evidence or updates the summary. However, in complex cases where historical and current information conflict, LC-Boost may struggle to fully resolve the issue relying solely on the accumulated summary and the current text.

% When processing data serially, if inconsistencies arise, CoA does not address the conflicts. Instead, it generates a summary based on the summaries of previous chunks and the current chunk, which is then passed on to the next worker. Similarly, LC-Boost does not resolve contradictions between different chunks but instead either appends new evidence (append action) or updates the summarization (merge action). Although LongAgent has the ability to handle conflicts, it requires concatenating the contexts of different members to regenerate the answer. This means that the model needs a longer context window, which also increases the time complexity. Additionally, the designs of CoA and LC-Boost are challenging to parallelize, requiring more time compared to the parallelizable MapReduce method in LangChain. 
% \comments{How CoA, LC-Boost address the challenge that different evidences lie across various chunks.}

In this paper, we introduce LLM$\times$MapReduce, a training-free framework for processing long texts that utilizes a divide-and-conquer approach, allowing models with short context windows to effectively handle long contexts.
To address the challenges of inter-chunk dependency and conflict, we introduce a {\em structured information protocol} and an {\em in-context confidence calibration} mechanism. The structured information protocol defines the information passed from the map stage to the reduce stage, ensuring the model has the critical inputs needed to infer the correct answer when aggregating different chunks. In-context confidence calibration allows the model to assign a reliable confidence score to the output of each chunk, aiding in effectively resolving inter-chunk conflicts. We evaluate the proposed method on various long-text benchmarks, and the experimental results show that our approach outperforms both closed- and open-source LLMs in terms of both performance and efficiency. 
Through ablation experiments, we further validate the effectiveness of each component in LLM$\times$MapReduce, demonstrating how each piece contributes to the overall performance.

\section{Approach}

% 1. Formal definition.
% 2. Workflow.
% 3. Figure.

\subsection{Problem Description}

In real-world scenarios, users may require the LLM to comprehend one or more lengthy documents that far exceed the model's effective context window. Formally, let $X$ represent the user-provided long text and $L$ denote the model's effective context length. In this work, we focus on cases where $\left | X \right | \gg L$, where $\left | X \right |$ represents the length of $X$. we partition the input text $X$ into a series of chunks $\{x_1, x_2, \cdots, x_n\}$, where the length of each chunk $x_i$ is within the model's effective context length $L$. For a given user query $Q$, the LLM, parameterized by $\bm{\theta}$, processes each chunk to generate intermediate outputs, which are then aggregated to predict the final answer.
% To formulate the task, we define the input for a long-text task as $Q $ and $ X$ , where  $Q$  represents the user's query and  $X$ is the long text. We partition  $X $ into multiple chunks  $\{x_1, x_2, \ldots, x_n\} $, where $n$ is the number of chunks . The process of obtaining the output using our framework can be expressed as $ y = \Phi(Q, X) $, where  $\Phi$ denotes the function that processes the query and long text to produce the final response  $y $.

\subsection{Workflow of LLM $\times$ MapReduce}

\begin{figure*}[t]
% \begin{center}
% %\framebox[4.0in]{$\;$}
% \fbox{\rule[-.5cm]{0cm}{4cm} \rule[-.5cm]{4cm}{0cm}}
% \end{center}
\centering
\includegraphics[width=0.98 \linewidth]{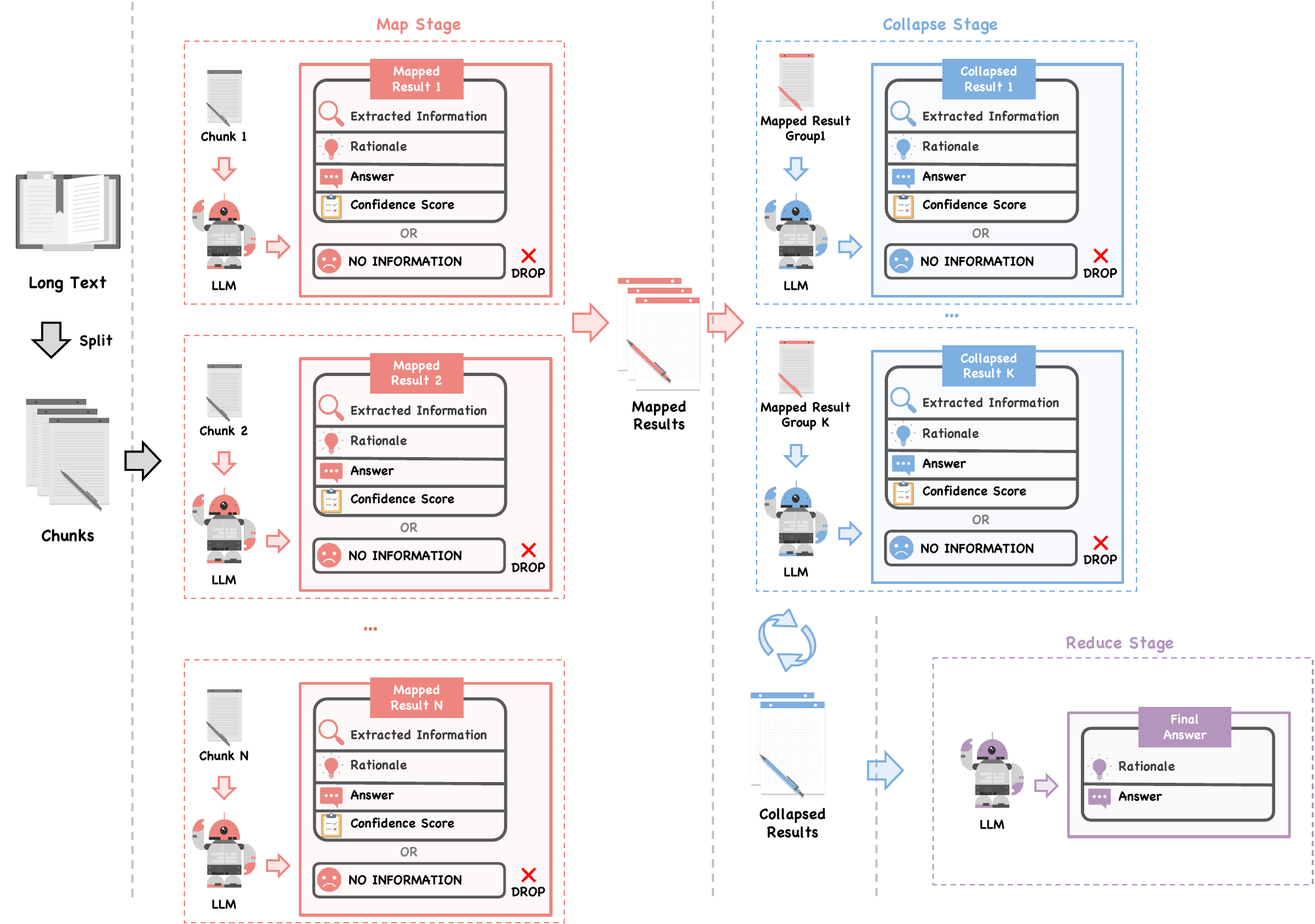}
\caption{Overview of the proposed LLM$\times$MapReduce framework. After dividing the provided long text into a series of chunks, the model processes each chunk to extract an information structure containing the essential content needed to address the query. This is referred to as the map stage in our framework. The mapped results are then compressed during the collapse stage, preparing them for the reduce stage. The structure of the collapsed results mirrors that of the mapped results. The collapse stage ensures that the input to the reducing model remains within its effective length (i.e., $L$). Based on the structured outputs from the first two stages (i.e., the map and collapse stages), the reduce model aggregates information from all chunks, resolves inter-chunk conflicts using calibrated confidence scores, and predicts the final answer.
% This framework follows a workflow similar to LangChain’s MapReduce method but with enhancements that enable the model to output results in a structured format. This structure facilitates the use of Chain-of-Thought (CoT, ~\citep{wei2023chainofthoughtpromptingelicitsreasoning}), allowing the model to more accurately extract relevant information and thoughtfully assess chunks that lack information. By doing so, the model effectively filters out outputs from chunks without information. Additionally, conflicts are resolved using the confidence scores within the structures.
}
\label{fig:workflow}
\end{figure*}

Figure~\ref{fig:workflow} depicts the overall framework of the proposed LLM$\times$MapReduce framework.
Like LangChain~\citep{langchain2022}, the LLM$\times$MapReduce workflow consists of three stages: map, collapse, and reduce.
During the map stage, we utilize an LLM as the map model to extract the necessary information for each chunk $x_i$:
\begin{equation}
    s_i = f_{\mathrm{map}}\left (x_i, Q;\bm{\theta} \right ),
\end{equation}
where $Q$ is the user query and $f_{\mathrm{map}}$ represents the map function powered by the LLM, parameterized by $\bm{\theta}$. Our experiments show that the design of the mapped results, $\{s_1, \cdots, s_N\}$, is crucial for enabling the divide-and-conquer framework to effectively comprehend long documents. 
In this work, we propose a structured information protocol aimed at improving communication efficiency between the different stages.
% which will be detailed in the following subsection.

In some cases, the input text is extremely long, resulting in mapped results that still exceed the context window of the LLM being used. To address this, a collapse stage is employed to compress the mapped results. We divide the $N$ mapped results into $K$ groups, ensuring that the length of each group remains within the model's context window $L$. For the $j$-th group of mapped results $g_j$, we leverage an LLM to output a compact result:
\begin{equation}
    c_j = f_{\mathrm{collapse}}\left (g_j, Q;\bm{\theta} \right ).
\end{equation}

It is important to note that the structure of each collapsed result $c_j$ remains the same as that of each mapped result $s_i$. 
If the total length of the mapped results $\{s_1, \cdots, s_N\}$ is less than $L$, we use the mapped results directly as the collapsed results for the reduce stage. If the collapsed results $\{c_1, \cdots, c_K\}$ still exceed $L$, we iteratively apply the collapse function $f_{\mathrm{collapse}}$ until their length is reduced to less than $L$. Briefly, we use $\{c_1, \cdots, c_K\}$ to denote the final output of the collapse stage.

% Our framework builds upon the MapReduce approach of LangChain and similarly includes three stages: map, collapse, and reduce. However, we introduce a specialized information structure to transfer information across different stages, along with an in-context confidence calibration mechanism to resolve conflicts that may arise between different chunks. 

% In the map stage, information is extracted from each chunk as  $s_i = \Phi(Q, x_i) $, where  $s_i $ represents the structured entity. The information structure $ s_i $ includes extracted information, rationale, answer, and confidence score. Through this process, we obtain a series of structured information entities $ S = \{ s_1, s_2, \ldots s_n\} $.  Subsequently, we filter out the structured entities that contain no information.

% Next, the length of the structured information $ S $ is assessed to determine if it exceeds the context window limit. If it does not, the collapse process is skipped. If the information exceeds the limit, a collapse process is applied to iteratively merge multiple pieces into a single entity that fits within the context window. This includes integrating extracted information, regenerating the rationale and answer, and assigning confidence scores, resulting in several structured information $ S' = \{s'_1, s'_2, \ldots s'_m\}$ fitting information structures.

Finally, in the reduce stage, the final response is generated based on the collapsed results:
\begin{equation}
    a = f_{\mathrm{reduce}}\left (\left\{ c_1, \cdots, c_K \right\}, Q;\bm{\theta} \right ).
\end{equation}
% information structures and user's query. In both the collapse and reduce stages, we guide the model to refer to the confidence scores within the structured entities to resolve conflicts between information from different chunks.
In LLM$\times$MapReduce, we do not need to tune the model parameters $\bm{\theta}$. Instead, the three functions (i.e., $f_{\mathrm{map}}$, $f_{\mathrm{collapse}}$, and $f_{\mathrm{reduce}}$) are implemented using prompts with existing LLMs.

The aforementioned divide-and-conquer framework is quite straightforward for long text processing, and has been explored in some previous studies~\citep{langchain2022,zhao2024longagentscalinglanguagemodels,zhang2024chainagentslargelanguage}. However, in our experiments, we find that simply combining an LLM and the divide-and-conquer strategy can not achieve satisfying performance on modern long-text benchmarks~\citep{zhang2024inftybench,hsieh2024ruler}. 

The major challenge is that segmenting the entire document may disrupt crucial long-range clues. The disrupted long-range information can be divided into two categories: inter-chunk dependency and inter-chunk conflicts. We therefore focus on enhancing the divide-and-conquer framework's ability to process cross-chunk information. 
Specifically, we propose a structured information protocol to address inter-chunk dependencies and an in-context confidence calibration mechanism to resolve inter-chunk conflicts. These approaches will be explained in the following subsections.

\subsection{Structured Information Protocol}

An important research question for divide-and-conquer long-text processing frameworks is determining {\em what information the map stage should convey to the reduce stage}.
If the mapped results are overly simplified, as seen in LongAgent~\citep{zhao2024longagentscalinglanguagemodels}, they may miss crucial details needed for subsequent stages (e.g., the reduce stage) to effectively handle inter-chunk dependencies and conflicts. On the other hand, if the mapped results are too complex, they introduce significant computational overhead, increasing the overall latency of the framework. Additionally, excessive unrelated information may interfere with the reduce model’s ability to produce the correct answer.

To this end, we introduce a specialized information structure consisting of four components:

\begin{itemize}
\item \textbf{Extracted Information}: key facts or data relevant to the query $Q$ that are extracted from the current chunk, providing the necessary background for subsequent stages to address inter-chunk dependencies.
\item \textbf{Rationale}: the analysis or inference process that explains how the model derives the intermediate answer from the extracted information, helping to mitigate the risk of hallucinations in subsequent stages.
\item \textbf{Answer}: the intermediate answer to the query, derived from the extracted information and rationale. If, after providing the rationale, the model determines that the passage does not contain relevant information to address the question, it will output ``\texttt{NO INFORMATION}'', which will be disregarded in subsequent stages.
\item \textbf{Confidence Score}: a score (out of 5) reflecting the model's confidence in the answer, indicating the completeness and reliability of the information. The confidence score is important for resolving inter-chunk conflicts.
\end{itemize}

To maintain a consistent input format for the reduce stage, both the map and collapse stages produce data in the structured format described above. ``Extracted Information'' together with ``Rationale'' provide the essential supplementary details needed for the reduce model to integrate answers from different chunks. Ablation experiments in Section~\ref{sec:ablation} demonstrate the effectiveness of the structured information protocol in enhancing the model's ability to manage inter-chunk dependencies.

A remaining issue with the structured information protocol is the potential inconsistency in confidence scores estimated across different chunks when resolving inter-chunk conflicts. Without a general criterion for confidence estimation, the model may assign varying levels of confidence to different chunks, even if the content is equally reliable. We thus propose an in-context confidence calibration mechanism to align the confidence scores of different chunks to a consistent standard.

% The use of CoT allows the model to extract relevant information more systematically, enabling it to carefully evaluate chunks that may lack content. This approach ensures the model accurately identifies and filters out non-informative chunks, reducing the risk of generating hallucinations. For QA tasks, the reduce stage is also designed to produce outputs in an information structure that includes both rationale and answer components. This structure employs the CoT mechanism to enhance the accuracy of the model's responses. 

% In different tasks, the content of the structure should be adjusted accordingly. For example, for general QA tasks, we employ the structure described previously. For code tasks, information extraction emphasizes retaining function signatures and brief comments while discarding irrelevant code. For summarization tasks, we streamline the process by requiring only a concise summary of text snippets and avoiding redundant responses. For retrieval tasks, confidence scores are constrained: direct evidence from the text receives a score of 5, while inferred evidence is capped at 5. For mathematical tasks, confidence scores place greater emphasis on the reasoning process. \comments{prompts , I think this paragraph can move to appendix }

\subsection{In-Context Confidence Calibration}

When long sequences are split into multiple chunks, conflicts may arise due to incomplete information. The aforementioned information structure provides a confidence score for the answer from each chunk. During the collapse and reduce stages, confidence scores are crucial for guiding the merging of information and generating both the rationale and the final answer.

To make the confidence scores across different chunks comparable, we propose to calibrate them through in-context learning, without adjusting model parameters. Specifically, we provide confidence estimation principles alongside a typical example for different levels of confidence score. By referring to the principles and the examples, the model is expected to apply a consistent criterion when processing different chunks. Figure~\ref{fig:in-context-example-prompt} provides an example of the calibration prompt. We can customize different calibration prompts for various tasks. Claims fully supported by the provided text are assigned high confidence, while those inferred by the model receive medium confidence. Claims not related to the provided text are assigned low confidence. Experiments in Section~\ref{sec:ablation} demonstrate the necessity of the proposed in-context confidence calibration mechanism.

\begin{figure}[h]
	\begin{tcolorbox}[colback=blue!2,colframe=blue!50!black]
	\small
	\texttt{Assign a confidence score (out of 5) to your answer based on the completeness and reliability of the extracted information and your rationale. The following is some assigning scoring cases:}\\
	\texttt{\textless Text: [}
	\textcolor{sblue}{\textbf{Jerry is 18 years old this year. He can swim and wants to be an athlete.}}
	\texttt{].\\Examples of confidence estimation: [}\\
	\textcolor{sgreen}{\textbf{Jerry can swim, 5 points;}}\\
	\textcolor{orange}{\textbf{Jerry will become an athlete in the future, 3.5 points; \\Jerry will become a swimming athlete in the future, 3 points;\\Jerry is strong, 3 points;}}\\
	\textcolor{sred}{\textbf{Jerry can play chess, 0 points;\\Jerry likes talking, 0 points}}
	\texttt{] \textgreater.}
	\end{tcolorbox}
	\caption{Prompt for in-context confidence calibration.}
	\label{fig:in-context-example-prompt}
	% \vspace{-1cm}
\end{figure}

\section{Experiments}

% 1. model, benchmark, baseline
% 2. results
\subsection{Setup}
\paragraph{Models}

We use two well-known open-source models to validate the effectiveness of the proposed LLM$\times$MapReduce framework, which are Llama3-70B-Instruct\footnote{\url{https://huggingface.co/meta-llama/Meta-Llama-3-70B-Instruct}} and Qwen2-72B-Instruct\footnote{\url{https://huggingface.co/Qwen/Qwen2-72B-Instruct}}. We employ vLLM\footnote{\url{https://github.com/vllm-project/vllm}} for model inference, and the decoding temperature is set to 0.7.

% Since our proposed MapReduce framework is model-agnostic, we can employ various LLMs in this framework. In our experiments, we utilized Meta-Llama-3-70B-Instruct ~\citep{llama3modelcard} for long sequences processing. The decoding parameters were set to a temperature of 0.7, top-p of 1.0, and top-k of -1. We employed vLLM \citep{kwon2023efficient} for model inference.
% In our study, we utilized the InfiniteBench benchmark to assess our long-text processing framework. InfiniteBench is the first benchmark for LLMs with an average data length exceeding 100K tokens. It includes comprehensive and real-world tasks across various domains, requiring the model to understand complex dependencies in long texts rather than merely retrieving text snippets. By conducting experiments on this benchmark, we aim to thoroughly evaluate and demonstrate the strengths and potential areas for improvement of our framework in handling long texts.

\paragraph{Evaluation}
We evaluate the performance of the involved models and methods on InfiniteBench~\citep{zhang2024inftybench}, where the average input length exceeds 100K tokens. This benchmark assesses the long-text capabilities of LLMs across several dimensions, including long-range retrieval, language comprehension, code understanding, and mathematical problem-solving. We exclude the subsets \texttt{Code.Run} and \texttt{Math.Calc}, as nearly all models achieve less than 5\% accuracy on these tasks, making it difficult to differentiate performance among the models. We utilize the evaluation code open-sourced by \citet{zhang2024inftybench} to calculate scores, with the exception of \texttt{En.Dia}. We find that the recall score for this task tends to increase with longer model outputs. Therefore, we directly engage two human experts with experience in natural language processing to manually assess the accuracy. 
% We utilized the InfiniteBench ~\citep{zhang2024inftybench} benchmark to evaluate our framework and other baselines. InfiniteBench is the first LLM benchmark with an average data length exceeding 100K tokens, requiring deep understanding of complex dependencies within long texts.  We follow the matrix used by InfiniteBench to evaluate the outputs of the LLM. For En.Dia task, since recall is used as the evaluation criterion and longer outputs receive higher scores, we directly employed two experts with natural language processing experience for manual evaluation. 
% Since Meta-Llama-3-70B-Instruct lacks Chinese capabilities, we evaluated all subsets of InfiniteBench except Zh.QA.
For \texttt{Retrieve.PassKey}, \texttt{Retrieve.Number}, \texttt{Retrieve.KV}, we use the retrieval prompt. For \texttt{En.Sum}, we use the summarization prompt. For \texttt{En.QA}, \texttt{En.MC}, \texttt{En.Dia}, we use the language question-answering prompt. For \texttt{Code.Debug}, we use the code prompt. For \texttt{Math.Find}, we use the math prompt.
% Please refer to the Appendix for the detailed content of the aforementioned prompt.

% and Math.Find tasks, we used retrieval prompts. For Code.Debug and Code.Run, we used code-specific prompts. For En.Sum, we used summarization prompts. For EM.QA, evaluated using the F1 score, we had the reduce phase output both the reasoning process and the answer, but only evaluated the answer. For En.Dia, evaluated by recall, we employed two NLP experts for manual evaluation. For the remaining tasks, we used the default prompts.

\paragraph{Baselines}

We select several representative models and methods as our baselines. For closed-source models, we compare against GPT-4, Claude 2~\citep{claude2}, and Kimi-Chat. For open-source models, we include YaRN-Mistral\footnote{\url{https://huggingface.co/NousResearch/Yarn-Mistral-7b-128k}}, Yi-6B-200K, Yi-34B-200K\footnote{\url{https://huggingface.co/01-ai}}, and Qwen2-72B-Instruct. Additionally, we compare LLM$\times$MapReduce with two recent representative frameworks for divide-and-conquer long-sequence processing: LongAgent~\citep{zhao2024longagentscalinglanguagemodels} and Chain-of-Agents~\citep{zhang2024chainagentslargelanguage}.

% The commercial models include GPT-4, Claude 2~\citep{claude2}, and Kimi-Chat\footnote{We present the performance of Kimi-Chat as reported in the InfiniteBench paper and our own test results as of 2024.06, considering potential updates to the Kimi-Chat.}, while the open-source models we selected are Qwen2-72B-Instruct, YaRN-Mistral-7B\footnote{\url{https://huggingface.co/NousResearch/Yarn-Mistral-7b-128k}, this model is used in InfiniteBench paper.}, Yi-6B-200K~\citep{Yi-6B-200K}, Yi-34B-200K~\citep{Yi-34B-200K}, and ChatGLM3-6B-128K~\citep{glm2024chatglm}. Additionally, we focused on models enhanced with specific text processing techniques to improve long-text handling capabilities, including the MapReduce approach implemented with LangChain based on Meta-Llama-3-70B-Instruct. We also assessed the impact of removing the in-context confidence calibration mechanism within the special structure in InfiniteBench. 

\subsection{Main Results}

\begin{table*}[t]
\centering
\small
\begin{tabular}{l|rrrrrrrrr|r}
    \toprule
    % Methods & Retrieve.PassKey & Retrieve.Number & Retrieve.KV & En.Sum & En.QA & En.MC & En.Dia & Code.Debug  & Math.Find  & Avg. \\
    \bf Methods & \bf Re.Pa & \bf Re.Nu & \bf Re.KV & \bf En.Sum & \bf En.QA & \bf En.MC & \bf En.Dia & \bf Co.De  & \bf Ma.Fi  & \bf Avg. \\
    \midrule
    \multicolumn{11}{c}{\em Closed-Source Models} \\
    \midrule
    GPT-4$^\star$ & \textbf{100.00} & \textbf{100.00} & 89.00 & 14.73 & 22.44 & 68.12 & 7.50 & 54.31  & 60.00  & 57.34 \\
    % Kimi-Chat$^\star$ & 98.14 & 95.42 & 53.60 & 17.96 & 16.52 & 72.49 & 11.00  & 17.77  & 12.57 & 43.94 \\
    Claude~2$^\star$ & 97.80 & 99.15 & 65.40 & 14.50 & 11.97 & 67.25 & 43.00  & 33.24  & 32.29   & 51.62\\
    Kimi-Chat & 99.32 & 97.45 & 69.20 & 29.94 & 18.81 & 79.91 & 15.50  & 38.32  & 18.57 & 51.89 \\
    \midrule
    \multicolumn{11}{c}{\em Open-Source Models} \\
    \midrule
    YaRN-Mistral$^\star$ & 92.71 & 58.31 & 0.00 & 9.09 & 9.55 & 29.26 & 4.50 & 23.60  & 17.14 & 27.13 \\
    Yi-6B-200K$^\star$ & \textbf{100.00} & 94.92 & 0.00 & 0.92 & 9.20 & 36.68 & 1.50  & 18.78  & 4.29   & 29.59 \\
    Yi-34B-200K$^\star$ & \textbf{100.00} & \textbf{100.00} & 0.00  & 1.33 & 12.17 & 46.29 & 3.50  & 21.32 & 25.71  & 34.48 \\
    % ChatGLM3-6B-128K & 92.20 & 80.68 & \textless 5 & \textless 5 & \textless 5 & 10.48 & \textless 5  & 7.36  & 7.71  & 22.65 \\
    Q2-72B-I & \textbf{100.00} & \textbf{100.00} & 29.00 & 31.85 & 21.97 & 81.66 & 23.00 & 45.43 & 59.71& 54.74 \\
    \midrule
    \multicolumn{11}{c}{\em Divide-and-Conquer Frameworks} \\
    \midrule
    L3-70B-I+LA & 99.32 &  93.05 &  74.60 & 2.19 &  \textbf{35.41} &  69.00 &  7.50 &  24.11  &  79.14  &  53.81\\
    L3-70B-I+CoA  & 9.32 &  15.59 &  1.80 & 10.10 &  7.03 &  27.51 &  9.50 &  18.27  &  44.57  &  15.97\\
    \cmidrule(lr){1-1} \cmidrule(lr){2-11}
    L3-70B-I$\times$MR  & \textbf{100.00}&  99.79 &  \textbf{98.89} & 30.63 &  34.70 & 82.10&  17.50 &  \textbf{62.94} &  \textbf{91.43} &  \textbf{68.66}\\
    Q2-72B-I$\times$MR  & \textbf{100.00} &  \textbf{100.00} &  98.80 & \textbf{32.39} & 23.13 & \textbf{83.84} &  \textbf{46.50} &  54.82  &  54.29 &  65.97\\
    \bottomrule
\end{tabular}
\caption{Results on InfiniteBench. ``$^\star$'' indicates that we directly use the model outputs released by \citet{zhang2024inftybench} and re-calculate the score. ``Q2-72B-I'' and ``L3-70B-I'' refer to Qwen2-72B-Instruct and Llama3-70B-Instruct, respectively. ``LA'' and ``CoA'' denote LongAgent~\citep{zhao2024longagentscalinglanguagemodels} and Chain-of-Agents~\citep{zhang2024chainagentslargelanguage}, which are two recent representative frameworks for divide-and-conquer long-sequence processing .}
\label{tab:main_result}
\end{table*}

% Table \ref{tab:main_result} shows the performance across various models or long-text processing methods. Our model, with confidence calibration and special information structure, achieved the highest scores in several task, with an average score of 68.64. 
 
% Specifically, it excels in retrieval tasks, with top results in Retrieve.PassKey, Retrieve.KV, and Math.Find. In question-answering tasks, it surpasses all other models in En.MC and En.QA, demonstrating superior accuracy and reliability.

% In the Code.Debug task, our model outperforms GPT-4 by a notable margin of 25.63 points, highlighting its effectiveness in debugging and code-related evaluations. For summarization tasks (En.Sum), our model's performance is second only to Qwen2-72B-Instruct, reflecting strong summarization capabilities. Additionally, our model achieves the highest average score of 68.64, outperforming all other models and methods included in the evaluation. This comprehensive performance underscores the effectiveness of our framework in delivering high accuracy across diverse tasks.

Table~\ref{tab:main_result} presents the performance of the involved methods on InfiniteBench. Among closed-source models, GPT-4 achieves the highest average score of 57.34. In the open-source category, Qwen2-72B-Instruct, which has a claimed effective length of 128K, achieves an average score of 54.74, even surpassing Claude 2 and Kimi-Chat.

For the divide-and-conquer methods, the backbone model used is Llama3-70B-Instruct, which has an effective context length of 8K, significantly shorter than the test examples in InfiniteBench. The results indicate that LongAgent~\citep{zhao2024longagentscalinglanguagemodels} outperforms CoA on nearly all subtasks. The average score of LongAgent is close to that of Qwen2-72B-Instruct (53.81 vs. 54.74). Surprisingly, the proposed LLM$\times$MapReduce method achieves the highest average score (68.66), outperforming both the closed-source models and the divide-and-conquer baselines.
Augmented by the proposed method, Llama3-70B-Instruct performs well on all the subtasks.
Additionally, our method is compatible with Qwen2-72B-Instruct, demonstrating its generalization capability.

% For the two divide-and-conquer baselines, LongAgent~\citep{zhao2024longagentscalinglanguagemodels} significantly outperforms CoA~\citep{zhang2024chainagentslargelanguage} (53.81 vs. 15.97)

% In comparison with both open-source and closed-source models, our framework demonstrates strong performance, especially with Qwen2-72B-Instruct using MapReduce, which achieves an average score of 65.97, and Llama3-70B-Instruct using MapReduce, leading with an average score of 68.66. 

% When evaluated against other divide-and-conquer frameworks, such as Chain-of-Agents and LongAgent, our LLM $\times$ MapReduce framework exhibit significantly better results, with Chain-of-Agents and LongAgent scoring 15.97 and 53.81, respectively. This highlights the effectiveness of our approach. We will further demonstrate the effectiveness of our approach in the next section.

\subsection{Ablation Study}
\label{sec:ablation}
% we compared the results with and without confidence calibration, as shown in Table \ref{tab:main_result}. We found that, without confidence calibration, our model showed slightly lower scores across nearly all tasks, except for QA.

% When compared to LangChain's model, which also employs a MapReduce approach but lacks both structured outputs and the use of confidence calibration, our model outperformed it across all tasks. This underscores the significant advantages of our approach in achieving superior performance through structured information processing and the application of confidence calibration.

As mentioned in Section~\ref{sec:intro}, the major challenge for divide-and-conquer long-sequence processing methods lies in addressing inter-chunk dependencies and resolving inter-chunk conflicts. In LLM$\times$MapReduce, we introduce a structured information protocol and an in-context confidence calibration mechanism, setting our method apart from existing divide-and-conquer baselines. We conduct ablation experiments to investigate the effect of the two components.

% Compared to other divide-and-conquer methods, our major contribution lies in the introduction of two mechanisms: the structured information protocol ("Struc.") and in-context confidence calibration ("Conf."). To further understand their contributions, we progressively remove these mechanisms and observe the model's performance through quantitative analysis in this section.

% As shown in Table~\ref{tab:ablation}, when removing the in-context confidence calibration mechanism, we observe a decrease in performance across all tasks, especially for English language understanding tasks (i.e., En.Av.). When both the confidence calibration and the structured information protocol are disabled, the performance drops more significantly compared to the full framework. For retrieval tasks (i.e., Re.Av.), there is a slight decline from 99.56 to 97.14. However, En.Av. suffers a much larger drop, decreasing by 15.30 points to 25.93. In Co.De., performance decreases substantially by 16.49 points. In Ma.Fi, the score falls sharply by 35.43 points. These results demonstrate the necessity of both the two proposed mechanisms.

As shown in Table~\ref{tab:ablation}, removing the in-context confidence calibration mechanism leads to a performance decline across all tasks, particularly in English language understanding tasks (i.e., En.Avg). When both confidence calibration and the structured information protocol are disabled, the performance drops even more significantly compared to the full framework. For Re.Avg, there is a slight decrease from 99.56 to 97.14, but En.Avg experiences a much larger drop, falling by 15.30 points to 25.93. In Co.De, performance decreases substantially by 16.49 points, and in Ma.Fi, the score drops sharply by 35.43 points. These results underscore the importance of both mechanisms in maintaining strong performance for long-sequence processing.

% The result highlights the importance of both the structured information protocol and confidence calibration. Together, they enhance the framework's effectiveness, each ensuring more robust and stable performance across tasks. 

\begin{table}[t]
\centering
\small
\begin{tabular}{l cccc}
    \toprule
    \bf Method & \bf Re.Avg & \bf En.Avg & \bf Co.De & \bf Ma.Fi \\
    \midrule
    L3-70B-I$\times$MR & 99.56 & 41.23 & 62.94 & 91.43 \\
    \cmidrule(lr){1-1} \cmidrule(lr){2-5}
    \quad -Conf. & 96.00 & 39.18 & 58.12 & 90.00 \\
    \quad \quad -Struc. & 97.14 & 25.93 & 46.45 & 56.00 \\
    \bottomrule
\end{tabular}
\caption{Effect of structured information protocol and in-context confidence calibration. ``Re.Avg'' and ``En.Avg'' denote the average performance on retrieval tasks and English language understanding tasks, respectively.}
\label{tab:ablation}
\end{table}

\subsection{Extremely Long Evaluation}

\begin{figure*}[ht]
    \centering 
    \includegraphics[width=0.99\linewidth]{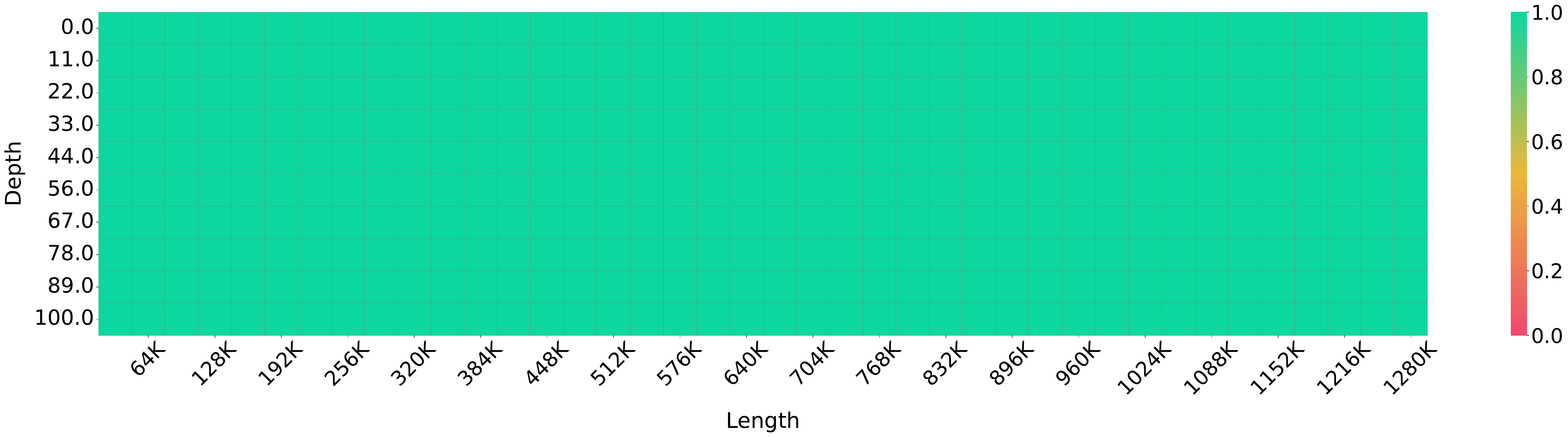} 
    \caption{Performance of  Llama3-70B-Instruct$\times$MapReduce on the NIAH test, with the maximum length of the haystack set to 1280K tokens.} 
    \label{fig:needle_in_a_haystack} 
\end{figure*}

Needle-in-a-haystack (NIAH)~\citep{niah} is a widely-used method for evaluating the ability of LLMs to handle long texts by identifying specific facts within long documents. To assess the performance of our framework in handling extremely long texts, we extend the NIAH test to a length of 1280K million tokens.   % needle in a haystack 的bib是我自己写的

Figure~\ref{fig:needle_in_a_haystack} presents the results, showing that our proposed method enables Llama3-70B-Instruct, which has a trained context length of 8K tokens, to effectively deal with sequences up to 1280K tokens. This demonstrates the potential of our framework for processing extremely long sequences.
% that, within the 1280k context, the Llama3-70B-Instruct $\times$ MapReduce framework maintains high accuracy in retrieving information across all text lengths and depths, demonstrating its robust information retrieval capabilities.

% \subsection{Evaluation on Ruler}

\subsection{Inference Latency}

\begin{figure}[h]
    \centering 
    \includegraphics[width=\linewidth]{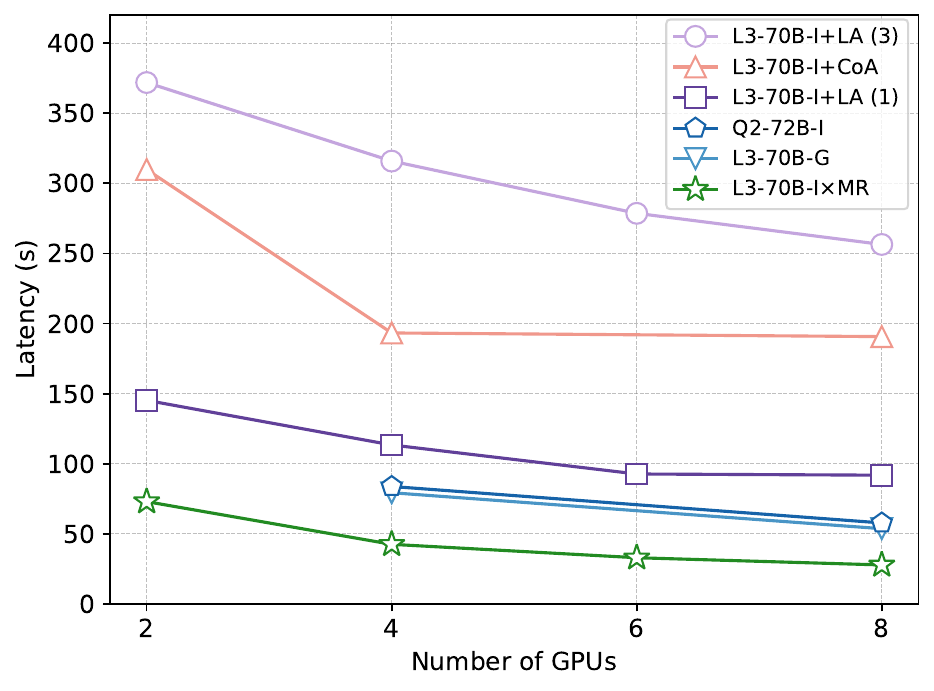} 
    \caption{Comparison of inference latency. ``L3-70B-G'' represents Llama3-70B-Instruct-Gradient-1048K.} % GPU怎么写还没改 
    \label{fig:speed} 
\end{figure}

Since divide-and-conquer long-sequence processing frameworks introduce multiple intermediate steps, they may be slower than standard decoding. To assess this, we measure the inference latency of the different approaches using 20 test examples, each with 128K tokens. Since the original Llama3-70B-Instruct does not support 128K tokens, we use Llama3-70B-Instruct-Gradient-1048K\footnote{\url{https://huggingface.co/gradientai/Llama-3-70B-Instruct-Gradient-1048k}}, an extended version of Llama3-70B-Instruct, to evaluate the inference speed. 
Since LongAgent uses a multi-turn mechanism to resolve inter-chunk conflicts, we report the latency for LongAgent with the maximum number of turns set to 1 and 3. The experiments are conducted using NVIDIA A100 GPUs (80 GB).

% To evaluate the inference speed of our framework, we conducted tests using 20 instances of the 128k Needle in a Haystack task. We compared the performance of our framework against other divide-and-conquer frameworks, including LongAgent and Chain-of-Agents, all of which are built upon Llama3-70B-Instruct as the base model. Additionally, we compared our framework with direct decoding using Llama-3-70B-Instruct-Gradient-1048k\footnote{\url{https://huggingface.co/gradientai/Llama-3-70B-Instruct-Gradient-1048k}} (GradientAI/Llama3) and Qwen2-72B-Instruct models. 

As shown in Figure~\ref{fig:speed}, both CoA and LongAgent are slower than standard decoding across different settings. However, a notable advantage of divide-and-conquer methods is their lower GPU requirements for handling long sequences. For standard decoding, at least 4 GPUs are needed to process 128K tokens, whereas divide-and-conquer methods can support 128K tokens using just 2 GPUs.
% Direct decoding was only evaluated for 4 and 8 GPU configurations because there is insufficient kv block memory for 2 GPUs and the model's attention heads cannot be evenly divided by 6 GPUs.
% The results, as shown in the Figure \ref{fig:speed}, demonstrate that our framework achieves competitive performance across different GPU configurations, consistently reducing inference time as the number of GPUs increases, and significantly surpassing other methods in several scenarios, particularly in multi-GPU settings.
Surprisingly, the proposed LLM$\times$MapReduce framework outperforms not only other divide-and-conquer baselines in speed but also standard decoding.
The efficiency of our method is achieved by avoiding the need to repeatedly process text chunks to resolve conflicts, as seen in LongAgent. Instead, we employ a structured information protocol and an in-context confidence calibration mechanism to effectively integrate information across chunks.

\section{Conclusion}

In this work, we introduce an effective divide-and-conquer framework for long-sequence processing, LLM$\times$MapReduce. With the structured information protocol and the in-context confidence calibration mechanism, LLM$\times$MapReduce effectively handles long texts, surpassing standard long-context LLMs and other divide-and-conquer baselines in terms of both effectiveness and efficiency.

% \section*{Acknowledgments}

\bibliography{custom}

% \appendix

% \section{Example Appendix}
% \label{sec:appendix}

% This is an appendix.

\end{document}